\documentclass{article} 
\usepackage{nips14submit_e,times}
\usepackage{bbfile}   
\usepackage{hyperref}
\usepackage{url}
\usepackage{mathtools}
\usepackage{nicefrac}
\usepackage{amsmath,amsfonts,amssymb}
\usepackage{pgfplots}
\usepackage{caption}
\usepackage{subcaption}
\usepackage[style=long,smallcaps]{glossaries}
 \newacronym
[longplural={Marginalised Gaussian Processes}]
{mgp}{mgp}{Marginalised Gaussian Process}
\newacronym
[longplural={Gaussian processes}]
{gp}{gp}{Gaussian process}
\newacronym{lda}{lda}{Latent Dirchlet}
\newacronym{mab}{mab}{Multi-Armed Bandit}
\newacronym{dpp}{dpp}{Determinantal Point Process}
\newacronym{abc}{abc}{Approximate Bayesian Computation}
\newacronym{eq}{eq}{Exponentiated Quadratic}
\newacronym{se}{se}{Squared Exponential}
\newacronym{nigp}{nigp}{Noisy Input Gaussian Process}
\newacronym{ard}{ard}{automatic relevance detection}
\newacronym{iid}{i.i.d.}{independently identically distributed}
\newacronym{ibcc}{ibcc}{Independent Bayesian Classifier Combination}
\newacronym{mle3}{mle-iii}{Type 3 Maximum Likelihood}
\newacronym{mle}{mle}{Maximum Likelihood Estimation}
\newacronym{mcmc}{mcmc}{Markov chain Monte Carlo}
\newacronym{skld}{skld}{Symmetric Kullback-Leibler Divergence}
\newacronym{gpgo}{gpgo}{Gaussian process global optimisation}
\newacronym{awgn}{awgn}{Additive White Gaussian Noise}
\newacronym{bq}{bq}{Bayesian quadrature}
\newacronym{nn}{nn}{Neural Network}
\newacronym{rq}{rq}{rational quadratic}
\newacronym{gps}{gps}{Global Positioning System}
\newacronym{mlhgp}{mlhgp}{Most Likely Heteroscedastic Gaussian Process}
\newacronym{glm}{glm}{Generalised Linear Model}
\newacronym{gpml}{gpml}{Gaussian Processes for Machine Learning Matlab Toolbox}
\newacronym{vb}{vb}{Variational Bayes}
\newacronym{psd}{psd}{positive semidefinite}
\newacronym{adc}{adc}{Analogue to Digital Converter}
\newacronym{dac}{dac}{Digital to Analogue Converter}
\newacronym{co}{co}{Carbon Monoxide}
\newacronym{ppm}{ppm}{Parts per Million}
\newacronym{icml}{icml}{International Conference on Machine Learning}
\newacronym{bo}{bo}{Bayesian optimisation}
\newacronym{uav}{uav}{Unmanned Aerial Vehicle}
\newacronym{hd}{hd}{High Definition}
\newacronym{ei}{ei}{Expected Improvement}
\newacronym{pi}{pi}{Probability of Improvement}
\newacronym{ucb}{ucb}{Upper Confidence Bound}
\newacronym{sdpp}{s-dpp}{Structure Determinantal Point Process}
\newacronym{soc}{soc}{State of Charge}
\newacronym{acfr}{acfr}{Australian Centre for Field Robotics}
\newacronym{uart}{uart}{Universal Asynchronous Receiver/Transmitter}
\newacronym{ndk}{ndk}{native development kit}
\newacronym{bbo}{bbo}{batch bayesian optimisation}
\newacronym{fitc}{fitc}{fully independent training conditional}
\newacronym{direct}{direct}{dividing rectangles}
\newacronym{el}{el}{expected loss}
\newacronym{rfr}{rfr}{random forest regressor}
\newacronym{nll}{nll}{negative log-likelihood}
\newacronym{ml}{ml}{machine learning}
\newacronym{go}{go}{global optimisation}
\newacronym{cmaes}{cma-es}{covariance matrix adaptive evolution strategy}
\newacronym{us}{wsabi}{warped sequential active Bayesian integration}
\newacronym{smc}{smc}{simple Monte Carlo}
\newacronym{bmc}{bmc}{Bayesian Monte Carlo}
\newacronym{ais}{ais}{annealed importance sampling}
\newacronym{bbq}{bbq}{Doubly-Bayesian Quadrature}

\usepackage{verbatim}
\usepackage{psfrag}
\usepackage[off]{auto-pst-pdf}
\usepackage{xspace}
\usepackage{hyperref}
\pgfplotsset{compat=newest}
\pgfplotsset{plot coordinates/math parser=false}
\usetikzlibrary{external}
\DeclareMathOperator*{\argmin}{arg\,min}
\DeclareMathOperator*{\argmax}{arg\,max}

    \renewcommand{\l}{\ell}
    \newcommand{\la}{\l_{\ast}}
    \newcommand{\xa}{x_{\ast}}
    \newcommand{\tl}{\tilde{\l}}
    \renewcommand{\v}[1]{\mathbf{#1}}
    
    \newcommand{\xd}{x_{d}}

    \newcommand{\momentsqrt}{\textsc{wsabi-m}\xspace}
    \newcommand{\linearsqrt}{\textsc{wsabi-l}\xspace}

    \newcommand{\lin}{\mathcal{L}}
    \newcommand{\mm}{\mathcal{M}}

    \newcommand{\N}{\mathcal{N}}
    \newcommand{\GP}{\mathcal{GP}}
    \newcommand{\D}{\mathcal{D}}

    \newcommand{\ud}{\mathrm{d}}

    \newcommand{\reals}{\mathbb{R}}

    \newcommand{\tCD}{\tilde{C}_{\D}}
    \newcommand{\tmD}{\tilde{m}_{\D}}
    \newcommand{\tlD}{\tmD}

    \makeatletter
    \newcommand*{\inlineequation}[2][]{%
    \begingroup
    \refstepcounter{equation}%
    \ifx\\#1\\%
    \else
      \label{#1}%
    \fi
    \relpenalty=10000 %
    \binoppenalty=10000 %
    \ensuremath{%
      #2%
    }%
    ~\@eqnnum
    \endgroup
    \makeatother
}

\title{Sampling for Inference in Probabilistic Models with Fast Bayesian Quadrature}

\author{
Tom Gunter, Michael A. Osborne\\
Engineering Science\\
University of Oxford\\
\texttt{\{tgunter,mosb\}@robots.ox.ac.uk} \\
\And
Roman Garnett \\
Knowledge Discovery and Machine Learning \\
University of Bonn \\
\texttt{rgarnett@uni-bonn.de} \\
\And
Philipp Hennig \\
MPI for Intelligent Systems \\
T\"ubingen, Germany \\
\texttt{phennig@tuebingen.mpg.de} \\
\And
Stephen J. Roberts \\
Engineering Science\\
University of Oxford\\
\texttt{sjrob@robots.ox.ac.uk} \\
}

\nipsfinalcopy 

\begin{document}

\maketitle

\begin{abstract}
  We propose a novel sampling framework for inference in probabilistic models: an active learning approach that converges more quickly (in wall-clock time) than Markov chain Monte Carlo (\textsc{mcmc}) benchmarks. The central challenge in probabilistic inference is numerical integration, to average over ensembles of models or unknown (hyper-)parameters (for example to compute the marginal likelihood or a partition function). \textsc{mcmc} has provided approaches to numerical integration that deliver state-of-the-art inference, but can suffer from sample inefficiency and poor convergence diagnostics. Bayesian quadrature techniques offer a model-based solution to such problems, but their uptake has been hindered by prohibitive computation costs. We introduce a warped model for probabilistic integrands (likelihoods) that are known to be non-negative, permitting a cheap active learning scheme to optimally select sample locations. Our algorithm is demonstrated to offer faster convergence (in seconds) relative to simple Monte Carlo and annealed importance sampling on both synthetic and real-world examples.
\end{abstract}

\section{Introduction}
Bayesian approaches to machine learning problems inevitably call for the frequent approximation of computationally intractable integrals of the form
\begin{equation}
Z = \langle \l \rangle = \int \l(\v{x})\,\pi(\v{x}) \, \ud\v{x},
\end{equation}
where both the likelihood $\l(\v{x})$ and prior $\pi(\v{x})$ are non-negative. Such integrals arise when marginalising over model parameters or variables, calculating predictive test likelihoods and computing model evidences. In all cases the function to be integrated---the integrand---is naturally constrained to be non-negative, as the functions being considered define probabilities.

In what follows we will primarily consider the computation of model evidence, $Z$. In this case $\l(\v{x})$ defines the unnormalised likelihood over a $D$-dimensional parameter set, $x_1,...,x_D$, and $\pi(\v{x})$ defines a prior density over $\v{x}$. Many techniques exist for estimating $Z$, such as \gls{ais} \cite{neal2001annealed}, nested sampling \cite{skilling2004nested}, and bridge sampling \cite{meng1996simulating}. These approaches are based around a core Monte Carlo estimator for the integral, and make minimal effort to exploit prior information about the likelihood surface. Monte Carlo convergence diagnostics are also unreliable for partition function estimates \cite{NealMC, brooks1998convergence, cowles1999possible}. More advanced methods---e.g., \gls{ais}---also require parameter tuning, and will yield poor estimates with misspecified parameters.

The \gls{bq} \cite{stanford1986bayesian, BZHermiteQuadrature, kennedy1998bayesian, BZMonteCarlo} approach to estimating model evidence is inherently model based. That is, it involves specifying a prior distribution over likelihood functions in the form of a \gls{gp} \cite{GPsBook}. This prior may be used to encode beliefs about the likelihood surface, such as smoothness or periodicity. Given a set of samples from $\l(\v{x})$, posteriors over both the integrand and the integral may in some cases be computed analytically (see below for discussion on other generalisations). Active sampling \citep{osborne2012active} can then be used to select function evaluations so as to maximise the reduction in entropy of either the integrand or integral. Such an approach has been demonstrated to improve sample efficiency, relative to na\"{i}ve randomised sampling \cite{osborne2012active}.

In a big-data setting, where likelihood function evaluations are prohibitively expensive, \gls{bq} is demonstrably better than Monte Carlo approaches \cite{BZMonteCarlo,osborne2012active}. As the cost of the likelihood decreases, however, \gls{bq} no longer achieves a higher effective sample rate per second, because the computational cost of maintaining the \gls{gp} model and active sampling becomes relevant, and many Monte Carlo samples may be generated for each new \gls{bq} sample. Our goal was to develop a cheap and accurate \gls{bq} model alongside an efficient active sampling scheme, such that even for low cost likelihoods \gls{bq} would be the scheme of choice. Our contributions extend existing work in two ways:

\textbf{Square-root \gls{gp}:}
Foundational work \cite{stanford1986bayesian, BZHermiteQuadrature, kennedy1998bayesian, BZMonteCarlo} on \gls{bq} employed a \gls{gp} prior directly on the likelihood function, making no attempt to enforce non-negativity a priori. \cite{osborne2012active} introduced an approximate means of modelling the logarithm of the integrand with a \gls{gp}. This involved making a first-order approximation to the exponential function, so as to maintain tractability of inference in the integrand model. In this work, we choose another classical transformation to preserve non-negativity---the square-root. By placing a \gls{gp} prior on the square-root of the integrand, we arrive at a model which both goes some way towards dealing with the high dynamic range of most likelihoods, and enforces non-negativity without the approximations resorted to in \cite{osborne2012active}.

\textbf{Fast Active Sampling:}
Whereas most approaches to \gls{bq} use either a randomised or fixed sampling scheme, \cite{osborne2012active} targeted the reduction in the expected variance of $Z$. Here, we sample where the expected posterior variance of the integrand after the quadratic transform is at a maximum. This is a cheap way of balancing exploitation of known probability mass and exploration of the space in order to approximately minimise the entropy of the integral.

We compare our approach, termed \emph{\gls{us}}, to non-negative integration with standard Monte Carlo techniques on simulated and real examples. Crucially, we make comparisons of error against ground truth \emph{given a fixed compute budget}.

\section{Bayesian Quadrature}
Given a non analytic integral $\langle \l \rangle \coloneqq \int \l(\v{x}) \pi(\v{x}) \, \ud\v{x}$ on a domain $\mathcal{X} = \mathbb{R}^D$, Bayesian quadrature is a model based approach of inferring both the functional form of the integrand and the value of the integral conditioned on a set of sample points. Typically the prior density is assumed to be a Gaussian, $\pi(\v{x}) \coloneqq \N(\v{x}; \boldsymbol{\nu}, \boldsymbol{\Lambda})$; however, via the use of an importance re-weighting trick, $q(\v{x}) = (\nicefrac{q(\v{x})}{\pi(\v{x})}) \, \pi(\v{x})$, any prior density $q(\v{x})$ may be integrated against. For clarity we will henceforth notationally consider only the $\mathcal{X} = \mathbb{R}$ case, although all results trivially extend to $\mathcal{X} = \reals^d$.

Typically a \gls{gp} prior is chosen for $\l(x)$, although it may also be directly specified on $\l(x)\pi(x)$. This is parameterised by a mean $\mu(x)$ and scaled Gaussian covariance
$K(x,x') \coloneqq \lambda^2 \exp \left (-\frac{1}{2}\frac{(x-x')^2}{\sigma^2} \right ).
$
The output length-scale $\lambda$ and input length-scale $\sigma$ control the standard deviation of the output and the autocorrelation range of each function evaluation respectively, and will be jointly denoted as $\theta = \{\lambda, \sigma\}$. Conditioned on samples $\xd = \{x_1,...,x_N\}$ and associated function values $\l(\xd)$, the posterior mean is
$m_{\D}(x) \coloneqq \mu(x) + K(x, \xd) K^{-1}(\xd,\xd) \bigl(\l(\xd)-\mu(\xd)\bigr)$,
and the posterior covariance is
$C_{\D}(x, x') \coloneqq K(x,x) - K(x,\xd) K(\xd,\xd)^{-1} K(\xd,x)$,
where $\D \coloneqq \bigl\{\xd,\l(\xd), \theta \bigr\}$. For an extensive review of the \gls{gp} literature and associated identities, see \citep{GPsBook}.

When a \gls{gp} prior is placed directly on the integrand in this manner, the posterior mean and variance of the integral can be derived analytically through the use of Gaussian identities, as in \cite{BZMonteCarlo}. This is because the integration is a linear projection of the function posterior onto $\pi(x)$, and joint Gaussianity is preserved through any arbitrary affine transformation. The mean and variance estimate of the integral are given as follows: \inlineequation[eq:bq_mean]{\mathbb{E}_{\l\mid \D}\bigl[\langle \l \rangle \bigr] = \int m_{\D}(x)\, \pi(x) \, \ud x}, and \inlineequation[eq:bq_var]{\mathbb{V}_{\l\mid \D}\bigl[\langle \l \rangle \bigr] = \iint C_{\D}(x, x')\, \pi(x) \, \ud x\, \pi(x') \, \ud x'}.
Both mean and variance are analytic when $\pi(x)$ is Gaussian, a mixture of Gaussians, or a polynomial (amongst other functional forms).

If the \gls{gp} prior is placed directly on the likelihood in the style of traditional Bayes--Hermite quadrature, the optimal point to add a sample (from an information gain perspective) is dependent only on $\xd$---the locations of the previously sampled points. This means that given a budget of $N$ samples, the most informative set of function evaluations is a design that can be pre-computed, completely uninfluenced by any information gleaned from function values \cite{minka2000dqr}. In \cite{osborne2012active}, where the log-likelihood is modelled by a \gls{gp}, a dependency is introduced between the uncertainty over the function at any point and the function value at that point. This means that the optimal sample placement is now directly influenced by the obtained function values.

\begin{figure}[h!]
\centering
\begin{subfigure}{.5\textwidth}
 \centering
    \input{./finalplots/bqfunconly1.tex}\includegraphics[width=1\textwidth, trim = 0mm 0mm 0mm 10mm, clip]{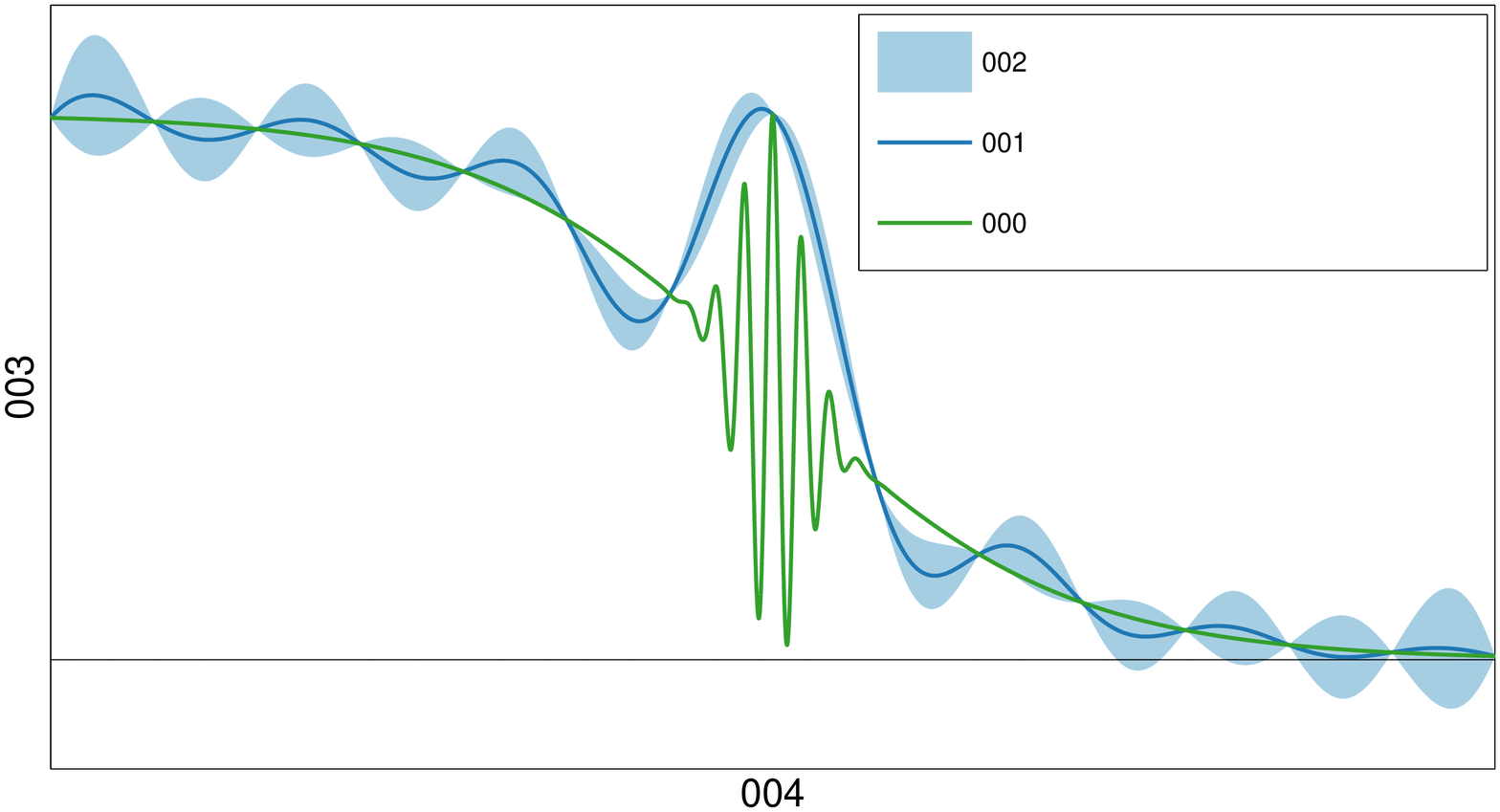}
    \captionof{figure}{Traditional Bayes--Hermite quadrature.}
    \label{fig:bayeshermite}
\end{subfigure}%
\begin{subfigure}{.5\textwidth}
  \centering
\input{./finalplots/momentsqrtfunconly1.tex}\includegraphics[width=1\textwidth, trim = 0mm 0mm 0mm 10mm, clip]{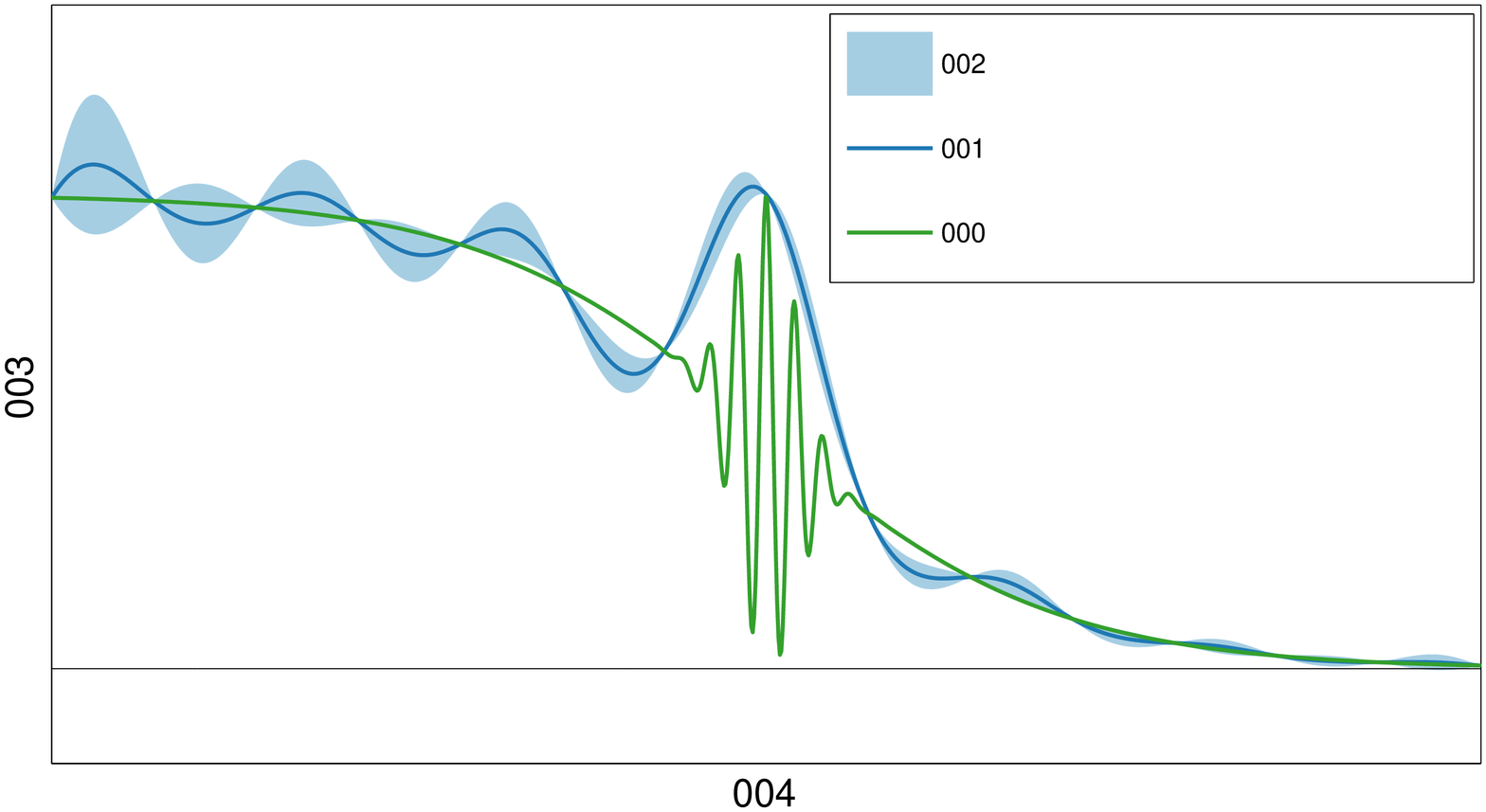}
\captionof{figure}{Square-root moment-matched Bayesian quadrature.}
\label{fig:sqrtbqgrid}
\end{subfigure}
\caption{Figure \ref{fig:bayeshermite} depicts the integrand as modelled directly by a \gls{gp}, conditioned on 15 samples selected on a grid over the domain. Figure \ref{fig:sqrtbqgrid} shows the moment matched approximation---note the larger relative posterior variance in areas where the function is high. The linearised square-root GP performed identically on this example, and is not shown.}
\label{fig:comparebqsqrtbq}
\end{figure}


An illustration of Bayes--Hermite quadrature is given in Figure \ref{fig:bayeshermite}. Conditioned on a grid of 15 samples, it is visible that any sample located equidistant from two others is equally informative in reducing our uncertainty about $\l(x)$. As the dimensionality of the space increases, exploration can be increasingly difficult due to the curse of dimensionality. A better designed \gls{bq} strategy would create a dependency structure between function value and informativeness of sample, in such a way as to appropriately express prior bias towards exploitation of existing probability mass.

\section{Square-Root Bayesian Quadrature}
Crucially, likelihoods are non-negative, a fact neglected by traditional Bayes--Hermite quadrature. In \cite{osborne2012active} the logarithm of the likelihood was modelled, and approximate the posterior of the integral, via a linearisation trick. We choose a different member of the power transform family---the square-root.

The square-root transform halves the dynamic range of the function we model. This helps deal with the large variations in likelihood observed in a typical model, and has the added benefit of extending the autocorrelation range (or the input length-scale) of the \gls{gp}, yielding improved predictive power when extrapolating away from existing sample points.

Let $\tl(x) \coloneqq \sqrt{2\bigl(\l(x)-\alpha\bigr)}$, such that $\l(x) = \alpha + \nicefrac{1}{2}\,\tl(x)^2$, where $\alpha$ is a small positive scalar.%
\footnote{%
   $\alpha$ was taken as $0.8 \times \min \l(\xd)$ in all experiments; our investigations found that performance was insensitive to the choice of this parameter.
}
We then take a \gls{gp} prior on $\tl(x)$: $\tl \sim \GP(0, K)$.
We can then write the posterior for
$\tl$ as
\begin{align}
    p(\tl\mid \D)
    & =
    \GP\bigl(\tl; \tilde{m}_{\D}(\cdot), \tilde{C}_{\D}(\cdot, \cdot)\bigr);
    \\
    \tmD(x)
    & \coloneqq
    K(x, \xd) K(\xd,\xd)^{-1}\tl(x_d);
    \\
    \tCD(x, x')
    & \coloneqq
    K(x,x') - K(x,\xd) K(\xd,\xd)^{-1} K(\xd,x').
\end{align}
The square-root transformation renders analysis intractable with this \gls{gp}: we arrive at a process whose marginal distribution for any $\l(x)$ is a non-central $\chi^2$ (with one degree of freedom). Given this process, the posterior for our integral is not closed-form. We now describe two alternative approximation schemes to resolve this problem.

\subsection{Linearisation}
We firstly consider a local linearisation of the transform $f\colon \tl \mapsto \l=\alpha+\nicefrac{1}{2}\,\tl^2$. As \gls{gp}s are closed under linear transformations, this linearisation will ensure that we arrive at a \gls{gp} for $\l$ given our existing \gls{gp} on $\tl$. Generically, if we linearise around $\tl_0$, we have $\l \simeq f(\tl_0) + f'(\tl_0) (\tl - \tl_0)$. Note that $f'(\tl) = \tl$: this simple gradient is a further motivation for our transform, as described further in Section \ref{sub:quadrature}. We choose $\tl_0 = \tlD$; this represents the mode of $p(\tl\mid\D)$. Hence we arrive at
\begin{equation}
    \l(x) \simeq \bigl(\alpha + \nicefrac{1}{2}\,\tlD(x)^2 \bigr)
    + \tlD(x)\, \bigl(\tl(x) - \tlD(x)\bigr)
    =
    \alpha -\nicefrac{1}{2}\,\tlD(x)^2 +\tlD(x)\,\tl(x)
    .
\label{eq:linearisation}
\end{equation}
Under this approximation, in which $\l$ is a simple affine transformation of $\tl$, we have
\begin{align}
    p(\l\mid \D)
    & \simeq
    \GP\bigl(\l; m_{\D}^{\lin}(\cdot), C_{\D}^{\lin}(\cdot, \cdot)\bigr);
    \\
    m_{\D}^{\lin}(x)
    & \coloneqq
    \alpha + \nicefrac{1}{2}\,\tlD(x)^2;
    \\
    C_{\D}^{\lin}(x, x')
    & \coloneqq
    \tlD(x)\tCD(x, x')\tlD(x').
\label{eq:lin_meancov}
\end{align}

\subsection{Moment Matching}
Alternatively, we consider a moment-matching approximation: $p(\l \mid \D)$ is approximated as a \gls{gp} with mean and covariance equal to those of the true $\chi^2$ (process) posterior. This gives
$p(\l \mid \D) \coloneqq
    \GP\bigl(\l; m^{\mm}_{\D}(\cdot), C^{\mm}_{\D}(\cdot, \cdot)\bigr),$
where
\begin{align}
    m_{\D}^{\mm}(x)
    & \coloneqq
    \alpha + \nicefrac{1}{2}\,\bigl(\tmD^2(x) + \tCD(x,x)\bigr);
\label{eq:lik_mean}
    \\
    C_{\D}^{\mm}(x, x')
    & \coloneqq
    \nicefrac{1}{2}\,\tCD(x, x')^2 + \tmD(x)\tCD(x, x')\tmD(x').
\label{eq:lik_cov}
\end{align}

\begin{figure}[h!]
\centering
\captionsetup{width=.90\textwidth}
\input{./finalplots/heatmap.tex}\includegraphics[width=1\textwidth, trim = 0mm 0mm 0mm 10mm, clip]{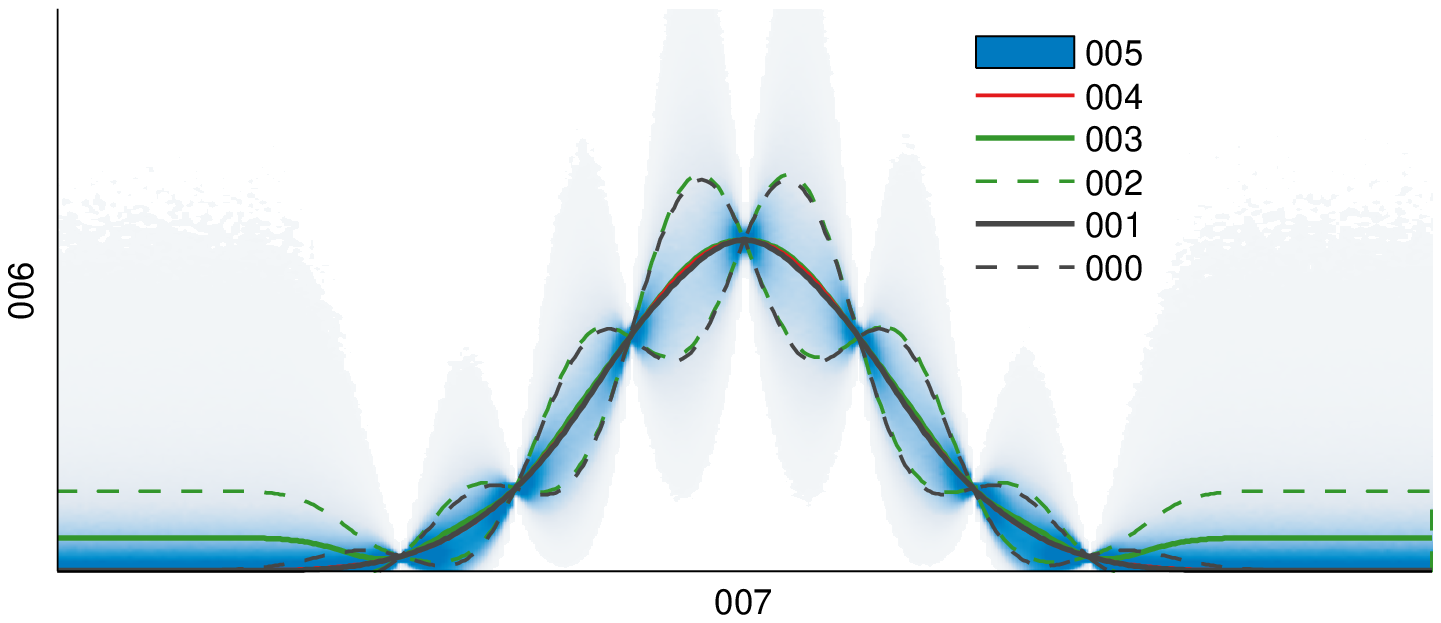}
\captionof{figure}{The $\chi^2$ process, alongside moment matched (\momentsqrt) and linearised approximations (\linearsqrt). Notice that the \linearsqrt mean is nearly identical to the ground truth.}
\label{fig:heatmap}
\end{figure}

We will call these two approximations \linearsqrt (for ``linear'') and
\momentsqrt (for ``moment matched''), respectively.  Figure
\ref{fig:heatmap} shows a comparison of the approximations on
synthetic data. The likelihood function, $\l(x)$, was defined to be
$\l(x) = \exp(-x^2)$, and is plotted in red. We placed a \gls{gp}
prior on $\tl$, and conditioned this on seven observations spanning
the interval $[-2, 2]$.  We then drew 50\,000 samples from the
true $\chi^2$ posterior on $\tl$ along a dense grid on the interval $[-5, 5]$ and
used these to estimate the true density of $\l(x)$, shown in blue
shading. Finally, we plot the means and 95\% confidence intervals for
the approximate posterior. Notice that the moment matching results in
a higher mean and variance far from observations, but otherwise the
approximations largely agree with each other and the true density.

\subsection{Quadrature} 
\label{sub:quadrature}

$\tmD$ and $\tCD$ are both mixtures of un-normalised Gaussians $K$. As such, the expressions for posterior mean and covariance under either the linearisation ($m_{\D}^{\lin}$ and $C_{\D}^{\lin}$, respectively) or the moment-matching approximations ($m_{\D}^{\mm}$ and $C_{\D}^{\mm}$, respectively) are also mixtures of un-normalised Gaussians. Substituting these expressions (under either approximation) into \eqref{eq:bq_mean} and \eqref{eq:bq_var} yields closed-form expressions (omitted due to their length) for the mean and variance of the integral $\langle \l \rangle$. This result motivated  our initial choice of transform: for linearisation, for example, it was only the fact that the gradient $f'(\tl)=\tl$ that rendered the covariance in \eqref{eq:lin_meancov} a mixture of un-normalised Gaussians. The discussion that follows is equally applicable to either approximation.

It is clear that the posterior variance of the likelihood model is now a function of both the expected value of the likelihood at that point, and the distance of that sample location from the rest of $\xd$. This is visualised in Figure \ref{fig:sqrtbqgrid}.

Comparing Figures \ref{fig:bayeshermite} and \ref{fig:sqrtbqgrid} we see that conditioned on an identical set of samples, \textsc{wsabi} both achieves a closer fit to the true underlying function, and associates minimal probability mass with negative function values. These are desirable properties when modelling likelihood functions---both arising from the use of the square-root transform.

\section{Active Sampling}
Given a full Bayesian model of the likelihood surface, it is natural to call on the framework of Bayesian decision theory, selecting the next function evaluation so as to optimally reduce our uncertainty about either the total integrand surface or the integral. Let us define this next sample location to be $\xa$, and the associated likelihood to be $\la \coloneqq \l(\xa)$. Two utility functions immediately present themselves as natural choices, which we consider below.  Both options are appropriate for either of the approximations to $p(\l)$ described above.

\subsection{Minimizing expected entropy}

One possibility would be to follow \cite{osborne2012active} in minimising the expected entropy of the integral, by selecting
$
    \xa
    =
    \argmin\limits_{x}
        \bigl\langle\mathbb{V}_{\l\mid\D,\ell(x)}
            \bigl[\langle \l \rangle \bigr]
        \bigr\rangle
$
, where
\begin{align}
    \Bigl\langle\mathbb{V}_{\l\mid\D,\ell(x)}
        \bigl[\langle \l \rangle \bigr]
    \Bigr\rangle
    & =
    \int \mathbb{V}_{\l\mid\D,\ell(x)}
        \bigl[\langle \l \rangle \bigr]
        \N \bigl(
            \ell(x); m_\D(x), C_\D(x, x) \,
        \bigr)
        \mathrm{d}\ell(x).
\label{eq:intentropy}
\end{align}

\subsection{Uncertainty sampling}

Alternatively, we can target the reduction in entropy of the total integrand $\l(x)\pi(x)$ instead, by targeting $\xa = \argmax\limits_{x} \mathbb{V}_{\l\mid\D} \bigl[\l(x)\pi(x)\bigr]$ (this is known as \emph{uncertainty sampling}), where
\begin{equation}
\mathbb{V}^{\mm}_{\l\mid\D} \bigl[\l(x)\pi(x)\bigr]
= \pi(x)C_{\D}(x, x)\pi(x)
= \pi(x)^2 \tCD(x, x)\bigl(\nicefrac{1}{2}\,\tCD(x, x) + \tmD(x)^2\bigr)
,
\label{eq:intvarMoment}
\end{equation}
in the case of our moment matched approximation, and, under the linearisation approximation,
\begin{equation}
\mathbb{V}^{\lin}_{\l\mid\D} \bigl[\l(x)\pi(x)\bigr]
= \pi(x)^2 \tCD(x, x)\tmD(x)^2.
\label{eq:intvarLinear}
\end{equation}

The uncertainty sampling option reduces the entropy of our \gls{gp} approximation to $p(\l)$ rather than the true (intractable) distribution. The computation of either \eqref{eq:intvarMoment} or \eqref{eq:intvarLinear} is considerably cheaper and more numerically stable than that of \eqref{eq:intentropy}. Notice that as our model builds in greater uncertainty in the likelihood  where it is high, it will naturally balance sampling in entirely unexplored regions against sampling in regions where the likelihood is expected to be high. Our model (the square-root transform) is more suited to the use of uncertainty sampling than the model taken in \cite{osborne2012active}. This is because the approximation to the posterior variance is typically poorer for the extreme log-transform than for the milder square-root transform. This means that, although the log-transform would achieve greater reduction in dynamic range than any power transform, it would also introduce the most error in approximating the posterior predictive variance of $\l(x)$. Hence, on balance, we consider the square-root transform superior for our sampling scheme.

\begin{figure}[h!]
\centering
\begin{minipage}{0.5\textwidth}
\captionsetup{width=.90\textwidth}
\input{./finalplots/momentsqrtbqactivesamples.tex}\includegraphics[width=1\textwidth, trim = 0mm 0mm 0mm 10mm, clip]{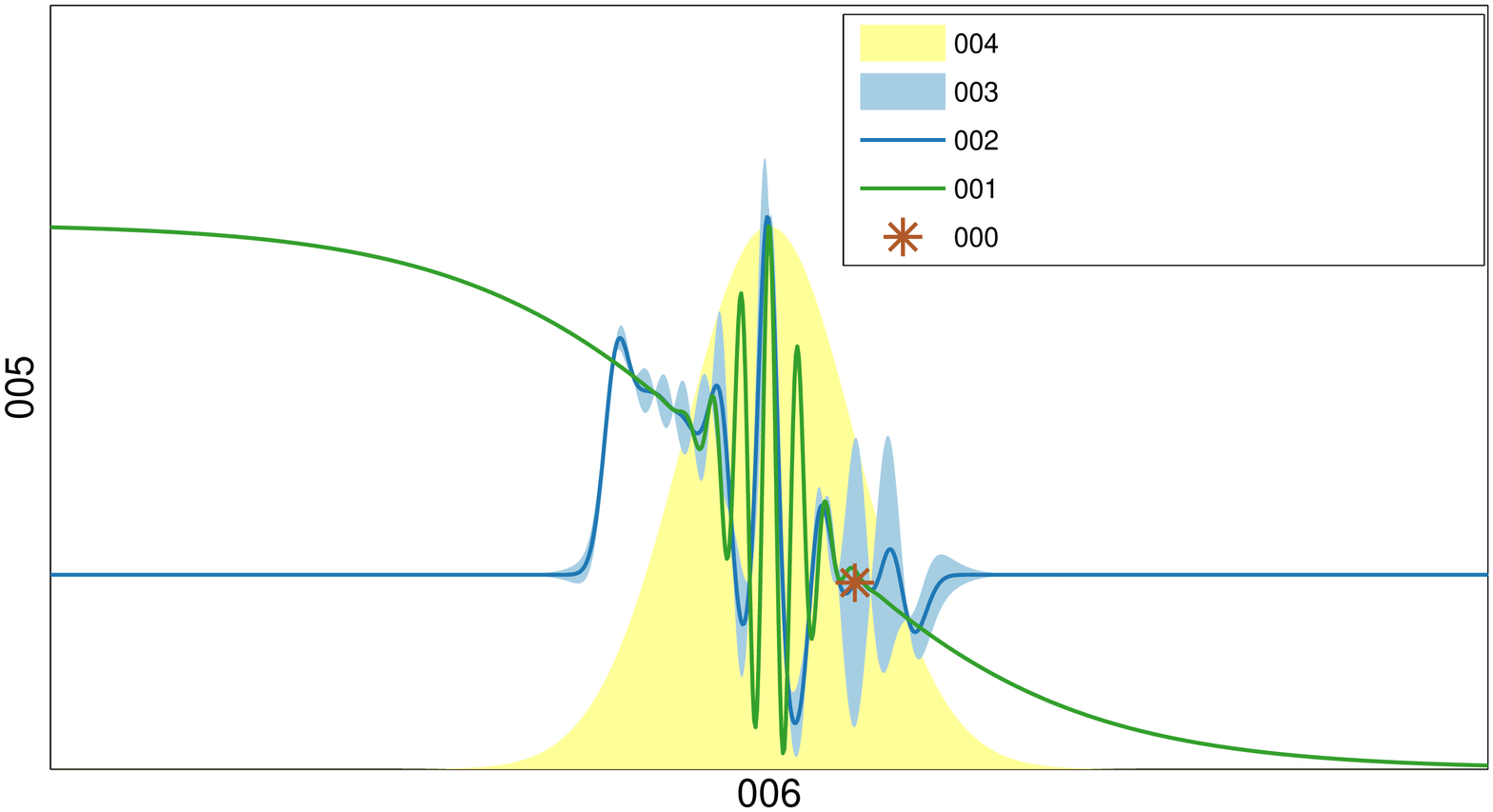}
\captionof{figure}{Square-root Bayesian quadrature with active sampling according to utility function \eqref{eq:intvarMoment} and corresponding moment-matched model. Note the non-zero expected mean everywhere.}
\label{fig:activesamplemoment}
\end{minipage}\hfill
\begin{minipage}{0.5\textwidth}
\centering
\captionsetup{width=.90\textwidth}
\input{./finalplots/linearsqrtbqactivesamples.tex}\includegraphics[width=1\textwidth, trim = 0mm 0mm 0mm 10mm, clip]{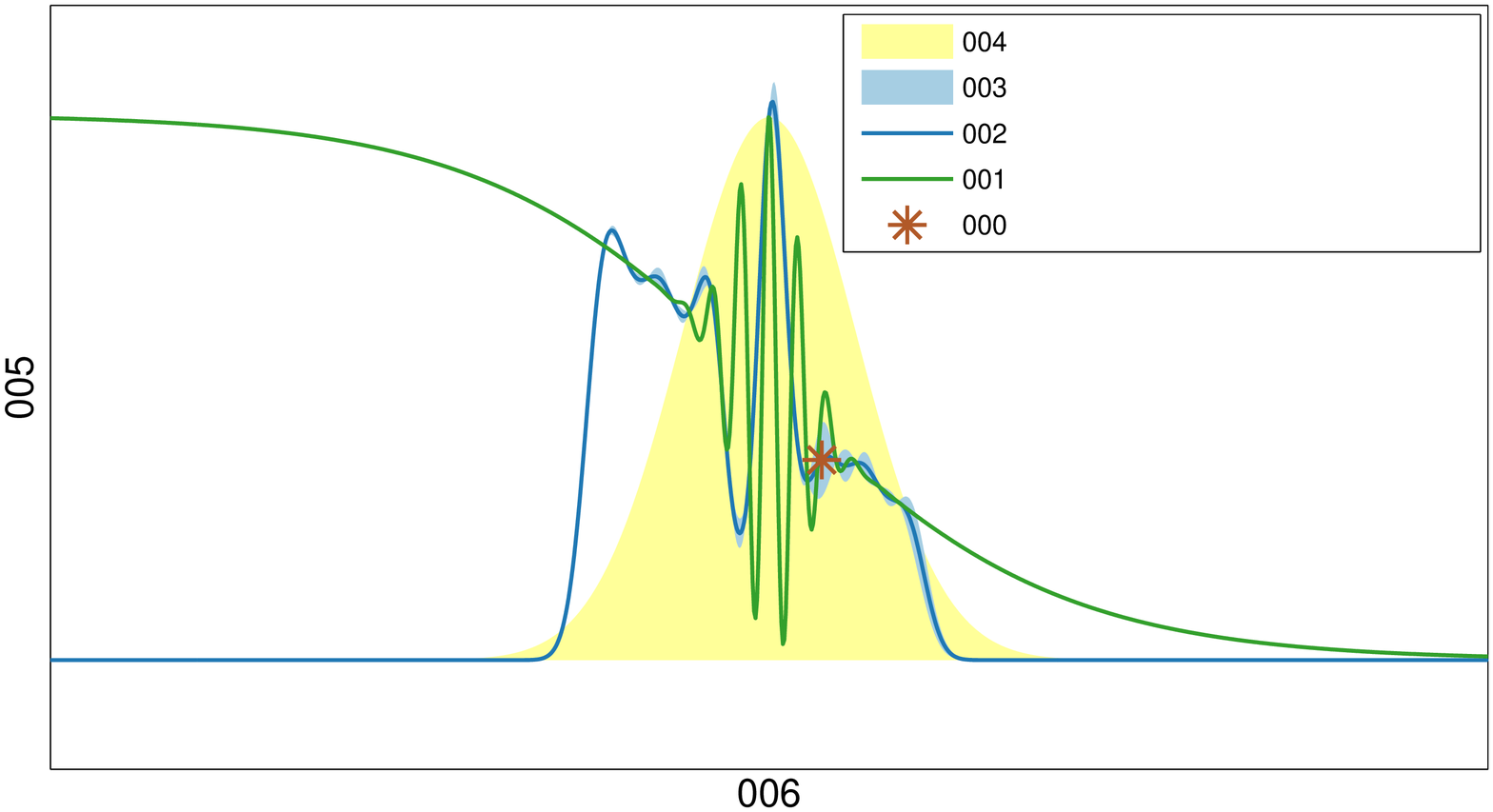}
\captionof{figure}{Square-root Bayesian quadrature with active sampling according to utility function \eqref{eq:intvarLinear} and corresponding linearised model. Note the zero expected mean away from samples.}
\label{fig:activesamplelinear}
\end{minipage}
\end{figure}

Figures \ref{fig:activesamplemoment}--\ref{fig:activesamplelinear} illustrate the result of square-root Bayesian quadrature, conditioned on 15 samples selected sequentially under utility functions \eqref{eq:intvarMoment} and \eqref{eq:intvarLinear} respectively. In both cases the posterior mean has not been scaled by the prior $\pi(x)$ (but the variance has). This is intended to exaggerate the contributions to the mean made by \momentsqrt.

A good posterior estimate of the integral has been achieved, and this set of samples is more informative than a grid under the utility function of minimising the integral error. In all active-learning examples a \gls{cmaes} \cite{hansen2003reducing} global optimiser was used to explore the utility function surface before selecting the next sample.

\section{Results}

Given this new model and fast active sampling scheme for likelihood surfaces, we now test for speed against standard Monte Carlo techniques on a variety of problems.

\subsection{Synthetic Likelihoods}

We generated 16 likelihoods in four-dimensional space by selecting $K$ normal distributions with $K$ drawn uniformly at random over the integers $5$--$14$. The means were drawn uniformly at random over the inner quarter of the domain (by area), and the covariances for each were produced by scaling each axis of an isotropic Gaussian by an integer drawn uniformly at random between $21$ and $29$. The overall likelihood surface was then given as a mixture of these distributions, with weights given by partitioning the unit interval into $K$ segments drawn uniformly at random---`stick-breaking'. This procedure was chosen in order to generate `lumpy' surfaces. We budgeted 500 samples for our new method per likelihood, allocating the same amount of time to \gls{smc}.

\begin{figure}[h!]
\centering
\begin{minipage}{.5\textwidth}
\captionsetup{width=.90\textwidth}
\centering
\input{./finalplots/gmmerrorlinear.tex}\includegraphics[width=1\textwidth, trim = 0mm 0mm 0mm 10mm, clip]{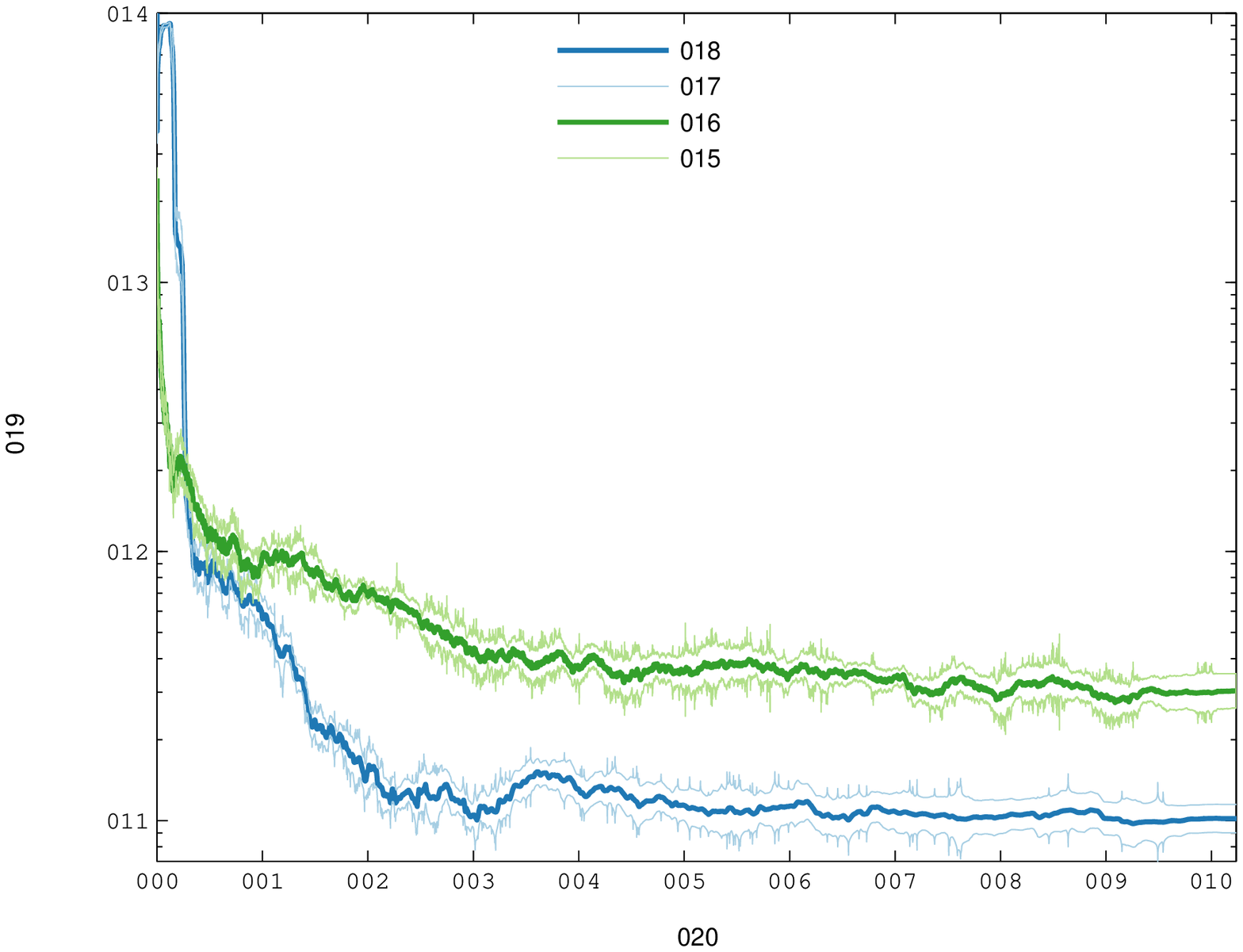}
\captionof{figure}{Time in seconds vs.\ average fractional error compared to the ground truth integral, as well as empirical standard error bounds, derived from the variance over the 16 runs. \momentsqrt performed slightly better.}
\label{fig:gmmerr}
\end{minipage}\hfill
\begin{minipage}{.5\textwidth}
\captionsetup{width=.90\textwidth}
\centering
\input{./finalplots/gmmlikelihoodlinear.tex}\includegraphics[width=1\textwidth, trim = 0mm 0mm 0mm 10mm, clip]{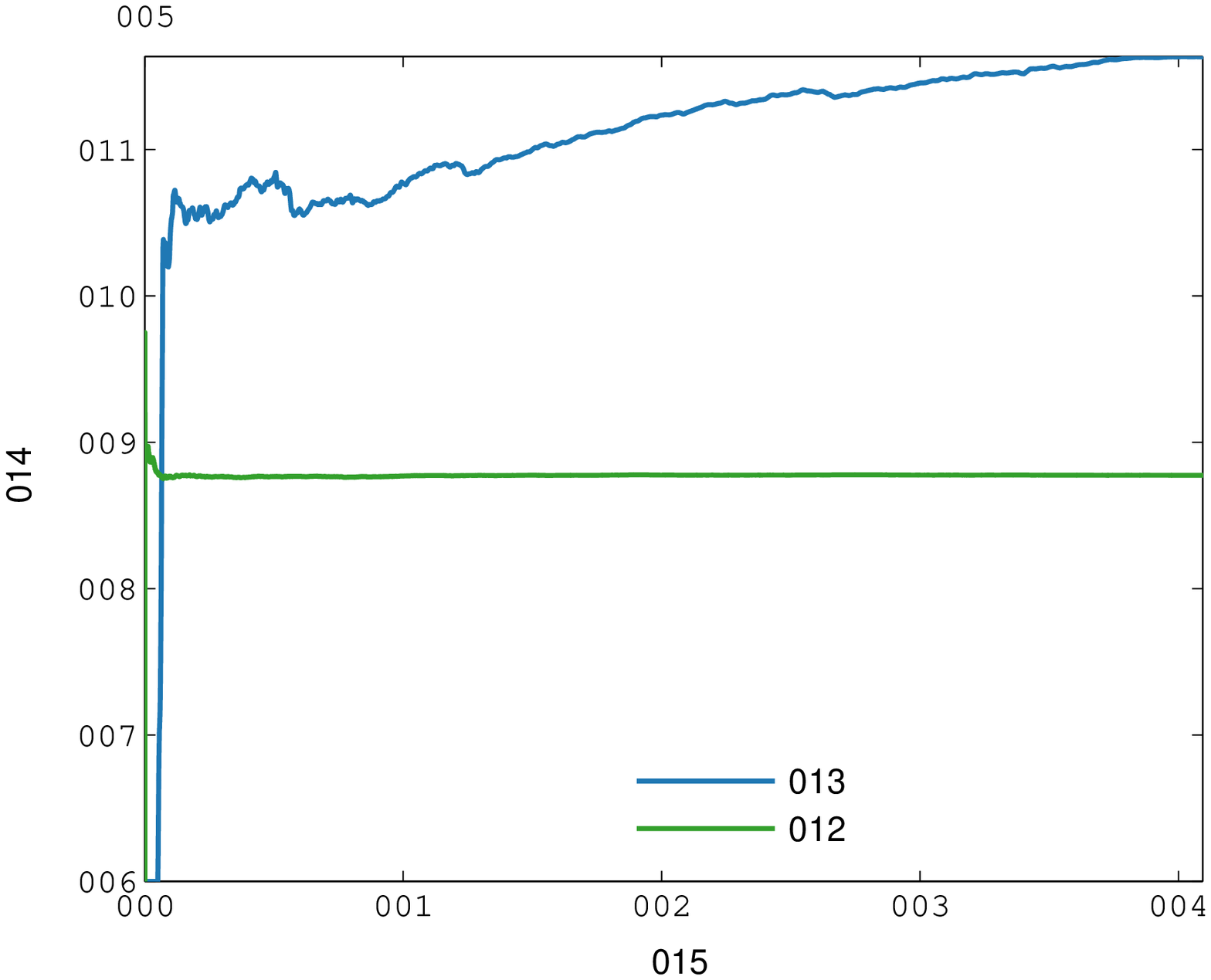}
\captionof{figure}{Time in seconds versus average likelihood of the ground truth integral over 16 runs. \momentsqrt has a significantly larger variance estimate for the integral as compared to \linearsqrt.}
\label{fig:gmmlik}
\end{minipage}
\end{figure}

Naturally the computational cost per evaluation of this likelihood is effectively zero, which afforded \gls{smc} just under 86\,000 samples per likelihood on average. \gls{us} was on average faster to converge to $10^{-3}$ error (Figure \ref{fig:gmmerr}), and it is visible in Figure \ref{fig:gmmlik} that the likelihood of the ground truth is larger under this model than with \gls{smc}. This concurs with the fact that a tighter bound was achieved.

\subsection{Marginal Likelihood of GP Regression}
As an initial exploration into the performance of our approach on real data,
we fitted a Gaussian process regression model to the \emph{yacht hydrodynamics} benchmark dataset \cite{yacht}. This has a six-dimensional input space corresponding to different properties of a boat hull, and a one-dimensional output corresponding to drag coefficient. The dataset has 308 examples, and using a squared exponential \textsc{ard} covariance function a single evaluation of the likelihood takes approximately 0.003 seconds.

Marginalising over the hyperparameters of this model is an eight-dimensional non-analytic integral. Specifically, the hyperparameters were: an output length-scale, six input length-scales, and an output noise variance. We used a zero-mean isotropic Gaussian prior over the hyperparameters in log space with variance of 4. We obtained ground truth through exhaustive \gls{smc} sampling, and budgeted 1\,250 samples for \gls{us}. The same amount of compute-time was then afforded to \gls{smc}, \gls{ais} (which was implemented with a Metropolis--Hastings sampler), and \gls{bmc}. \gls{smc} achieved approximately 375\,000 samples in the same amount of time. We ran \gls{ais} in 10 steps, spaced on a log-scale over the number of iterations, hence the \gls{ais} plot is more granular than the others (and does not begin at 0). The `hottest' proposal distribution for \gls{ais} was a Gaussian centered on the prior mean, with variance tuned down from a maximum of the prior variance.

\begin{figure}[h!]
\centering
\input{./finalplots/yachtresults1.tex}\includegraphics[width=0.8\textwidth, trim = 0mm 10mm 0mm 10mm, clip]{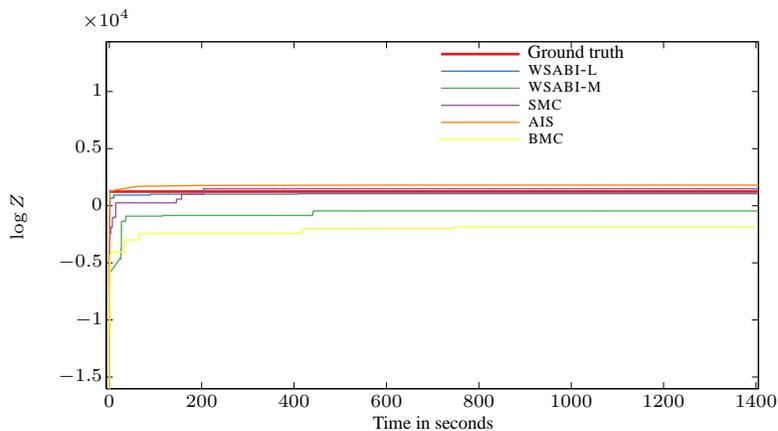}
\caption[]{Log-marginal likelihood of \gls{gp} regression on the yacht hydrodynamics dataset.}
\label{fig:marglik}
\end{figure}

Figure \ref{fig:marglik} shows the speed with which \gls{us} converges to a value very near ground truth compared to the rest. \gls{ais} performs rather disappointingly on this problem, despite our best attempts to tune the proposal distribution to achieve higher acceptance rates.

Although the first datapoint (after 10\,000 samples) is the second best performer after \gls{us}, further compute budget did very little to improve the final \gls{ais} estimate. \gls{bmc} is by far the worst performer. This is because it has relatively few samples compared to \gls{smc}, and those samples were selected completely at random over the domain. It also uses a \gls{gp} prior directly on the likelihood, which due to the large dynamic range will have a poor predictive performance.

\subsection{Marginal Likelihood of GP Classification}
We fitted a Gaussian process classification model to both a one dimensional synthetic dataset, as well as real-world binary classification problem defined on the nodes of a citation network \cite{garnett_et_al_icml_2012}. The latter had a four-dimensional input space and 500 examples. We use a probit likelihood model, inferring the function values using a Laplace approximation. Once again we marginalised out the hyperparameters.

\subsection{Synthetic Binary Classification Problem}
We generate 500 binary class samples using a 1D input space. The \gls{gp} classification scheme implemented in \gls{gpml} \citep{rasmussen2010gaussian} is employed using the inference and likelihood framework described above. We marginalised over the three-dimensional hyperparameter space of: an output length-scale, an input length-scale and a `jitter' parameter. We again tested against \gls{bmc}, \gls{ais}, \gls{smc} and, additionally, \gls{bbq} \citep{osborne2012active}. Ground truth was found through 100\,000 \gls{smc} samples.

\begin{figure}[h!]
\centering
\begin{minipage}{.5\textwidth}
\captionsetup{width=.90\textwidth}
\centering
\input{./finalplots/syntheticresults2.tex}\includegraphics[width=1\textwidth, trim = 0mm 0mm 0mm 10mm, clip]{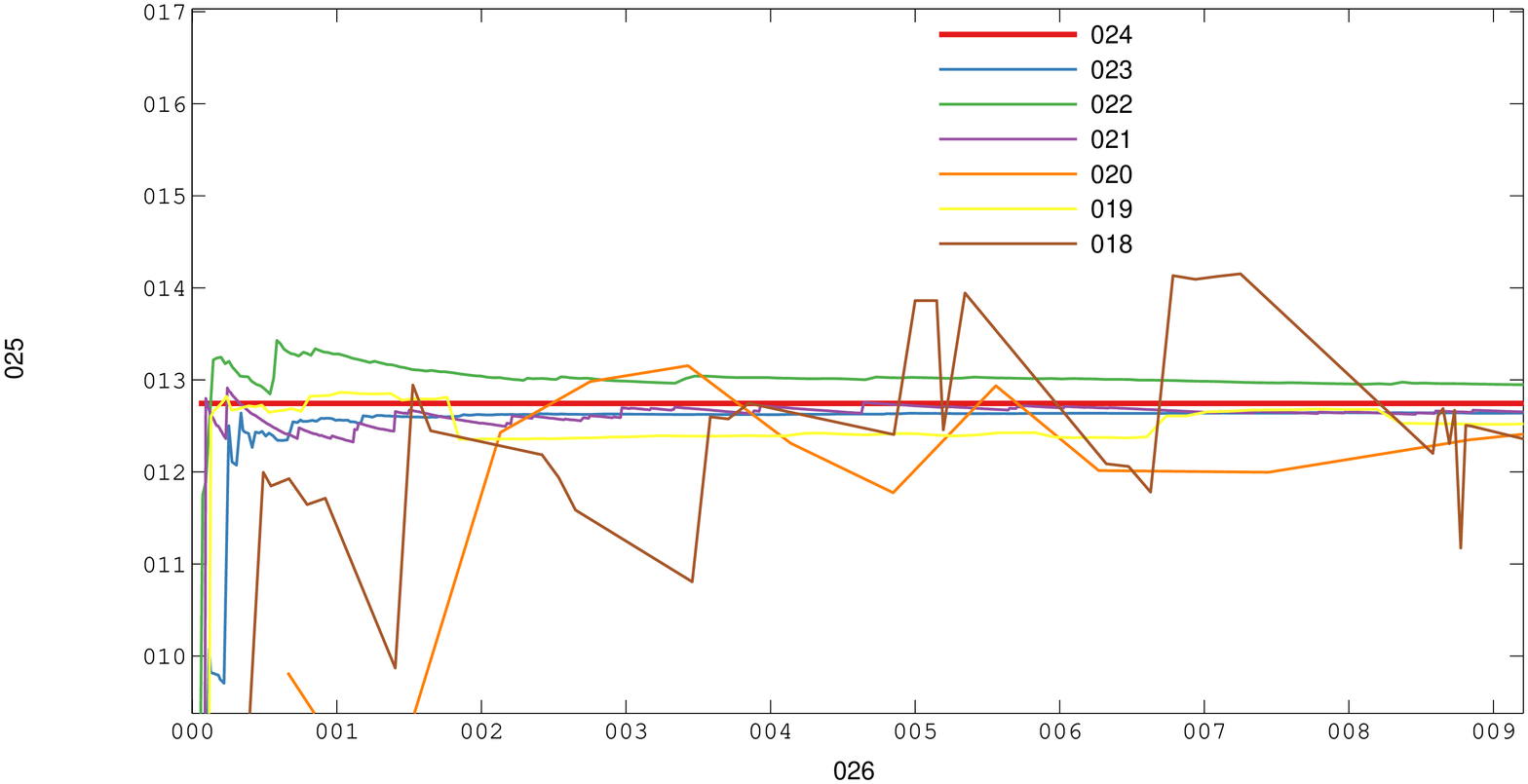}
\captionof{figure}{Log-marginal likelihood for \gls{gp} classification---synthetic dataset.}
\label{fig:synthclass}
\end{minipage}\hfill
\begin{minipage}{.5\textwidth}
\captionsetup{width=.90\textwidth}
\centering
\input{./finalplots/nipsclass.tex}\includegraphics[width=1\textwidth, trim = 0mm 0mm 0mm 10mm, clip]{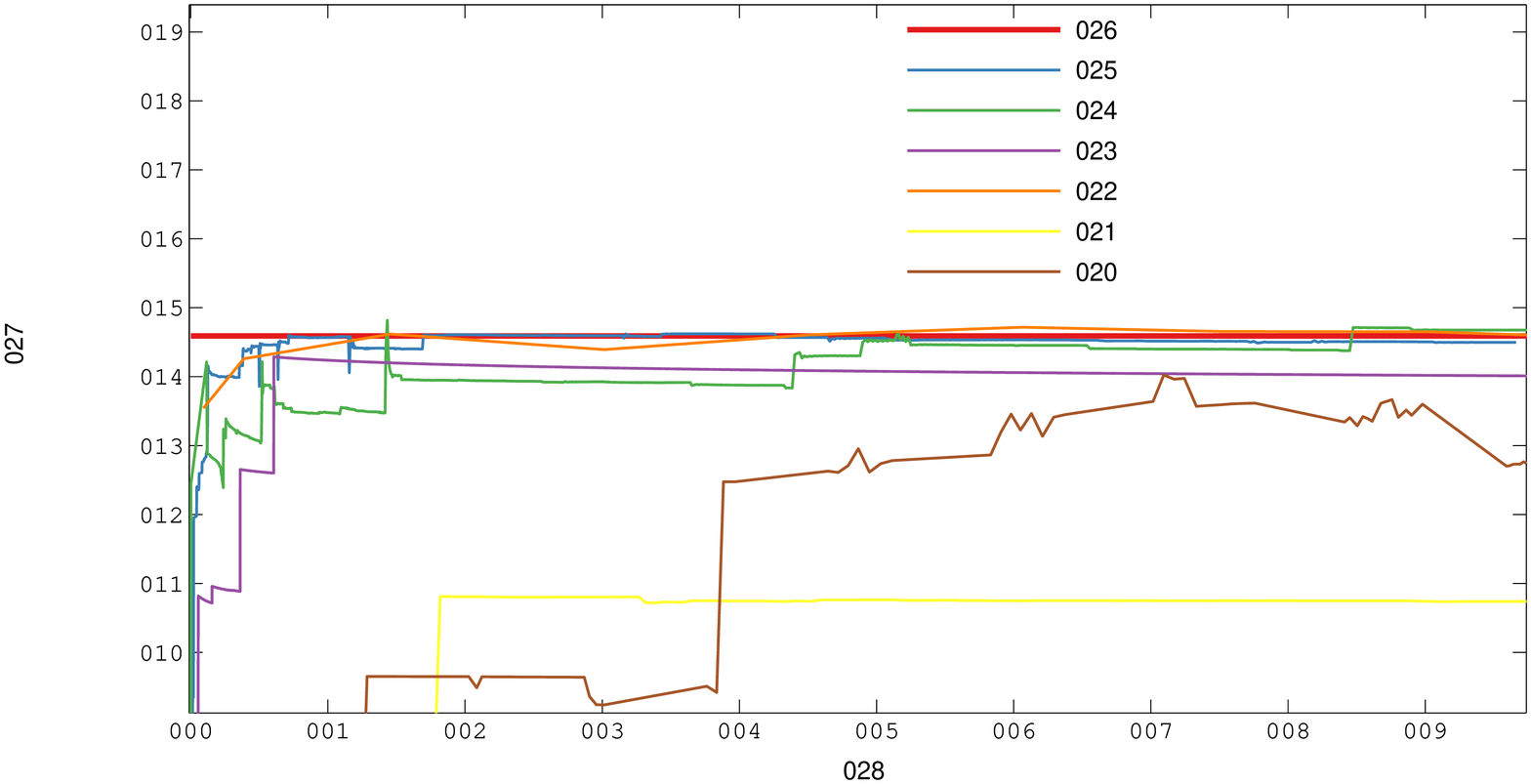}
\captionof{figure}{Log-marginal likelihood for \gls{gp} classification---graph dataset.}
\label{fig:realclass}
\end{minipage}
\end{figure}

This time the acceptance rate for \gls{ais} was significantly higher, and it is visibly converging to the ground truth in Figure \ref{fig:synthclass}, albeit in a more noisy fashion than the rest. \linearsqrt performed particularly well, almost immediately converging to the ground truth, and reaching a tighter bound than \gls{smc} in the long run. \gls{bmc} performed well on this particular example, suggesting that the active sampling approach did not buy many gains on this occasion. Despite this, the square-root approaches both converged to a more accurate solution with lower variance than \gls{bmc}. This suggests that the square-root transform model generates significant added value, even without an active sampling scheme. The computational cost of selecting samples under \gls{bbq} prevents rapid convergence.

\subsection{Real Binary Classification Problem}
For our next experiment, we again used our method to calculate the model evidence of a \gls{gp} model with a probit likelihood, this time on a real dataset.

The dataset, first described in \citep{garnett_et_al_icml_2012}, was a
graph from a subset of the CiteSeer\textsuperscript{x}\,citation
network. Papers in the database were grouped based on their venue of
publication, and papers from the 48 venues with the most associated
publications were retained.  The graph was defined by having these
papers as its nodes and undirected citation relations as its edges. We
designated all papers appearing in \textsc{nips} proceedings as
positive observations. To generate Euclidean input vectors, the
authors performed ``graph principal component analysis'' on this
network \citep{fouss}; here, we used the first four graph principal
components as inputs to a \gls{gp} classifier. The dataset was
subsampled down to a set of 500 examples in order to generate a cheap
likelihood, half of which were positive.

Across all our results, it is noticeable that \momentsqrt typically performs worse relative to \linearsqrt as the dimensionality of the problem increases. This is due to an increased propensity for exploration as compared to \linearsqrt. \linearsqrt is the fastest method to converge on all test cases, apart from the synthetic mixture model surfaces where \momentsqrt performed slightly better (although this was not shown in Figure \ref{fig:gmmerr}). These results suggest that an active-sampling policy which aggressively exploits areas of probability mass before exploring further afield may be the most appropriate approach to Bayesian quadrature for real likelihoods.

\section{Conclusions} 
\label{sec:conclusions}

We introduced the first fast Bayesian quadrature scheme, using a novel warped likelihood model and a novel active sampling scheme. Our method, \gls{us}, demonstrates faster convergence (in wall-clock time) for regression and classification benchmarks than the Monte Carlo state-of-the-art.

\pagebreak
\bibliography{bub}
\bibliographystyle{unsrt}

\end{document}